\documentclass{article}
\pdfoutput=1

\usepackage{microtype}
\usepackage{graphicx}
\usepackage{subfigure}
\usepackage{booktabs} %
\usepackage{times}
\usepackage{epsfig}
\usepackage{graphicx}
\usepackage{array,multirow}
\usepackage{adjustbox}
\usepackage[graphicx]{realboxes}
\usepackage{caption}
\usepackage{float}
\usepackage{amsmath}
\usepackage{dblfloatfix}
\DeclareMathOperator*{\argmax}{\arg\!\max}

\usepackage{amssymb}
\usepackage{bm}

\usepackage{hyperref}
\usepackage[capitalise, noabbrev]{cleveref}

\usepackage[colorinlistoftodos, textsize=tiny]{todonotes}
\definecolor{citrine}{rgb}{0.89, 0.82, 0.04}
\definecolor{blued}{RGB}{70,197,221}

\usepackage[accepted]{icml2019}

\icmltitlerunning{Adversarial Pixel-Level Generation of Semantic Images}

\begin{document}

\twocolumn[
	\icmltitle{Adversarial Pixel-Level Generation of Semantic Images}

	\icmlsetsymbol{equal}{*}

	\begin{icmlauthorlist}
		\icmlauthor{Emanuele Ghelfi}{to}
		\icmlauthor{Paolo Galeone}{to}
		\icmlauthor{Michele De Simoni}{to}
		\icmlauthor{Federico Di Mattia}{to}
	\end{icmlauthorlist}

	\icmlaffiliation{to}{Zuru Tech, Modena, Italy}

	\icmlcorrespondingauthor{Emanuele Ghelfi}{emanuele@zuru.tech}

	\icmlkeywords{Machine Learning, ICML, Generative Adversarial Network, Semantic}

	\vskip 0.3in
]

\printAffiliationsAndNotice{}  %

\begin{abstract}

	Generative Adversarial Networks (GANs) have obtained extraordinary success in the generation of realistic images, a domain where a lower pixel-level accuracy is acceptable. We study the problem, not yet tackled in the literature, of generating semantic images starting from a prior distribution. Intuitively this problem can be approached using standard methods and architectures. However, a better-suited approach is needed to avoid generating blurry, hallucinated and thus unusable images since tasks like semantic segmentation require pixel-level exactness. In this work, we present a novel architecture for learning to generate pixel-level accurate semantic images, namely Semantic Generative Adversarial Networks (SemGANs). The experimental evaluation shows that our architecture outperforms standard ones from both a quantitative and a qualitative point of view in many semantic image generation tasks.

\end{abstract}

\section{Introduction}

Generative Adversarial Networks \cite{goodfellow-generative-2014} (GANs) are generative methods used for estimating the relationship between simple latent distributions to complex distributions. Recent advances over the original GANs formulation involve the architecture \cite{radford-unsupervised-2015, mirza-conditional-2014, donahue-adversarial-2016, donahue-semantically-2018}, the training technique \cite{salimans-improved-2016}, or the applications  \cite{luc-semantic-2016, zhu-unpaired-2017}.

In this paper, we address the problem, not yet tackled, of generating semantic images starting from a latent representation. In this context, a semantic image is an image in which every pixel is assigned to a label indicating its class, thus assuming only values in a discrete and finite set. We propose a new architecture able to generate semantic images with higher precision than a standard architecture.

While the image-to-image approach \citep{isolaImagetoImageTranslationConditional2016} has proven to be effective in this domain, we argue that in this case, it is not possible to generate completely new images starting from an arbitrary latent representation, as an input image is always required. This limitation makes an image-to-image approach less flexible w.r.t. our approach since it cannot leverage the power of a learned latent, low-dimensional, representation. Moreover, the presented approach is also faster not requiring an encoder-like network embedded in the generator.

The contributions of this paper are the followings:
\begin{itemize}
	\item we present a new set of applications for the GAN framework: the generation of semantic images from latent vectors;
	\item we propose, to the best of our knowledge, the first GAN architecture that generates semantic images directly, without the need of any post-process step;
    \item the experimental evaluation of the new architecture on several semantic datasets; showing that our approach leads to improved image quality when compared to the standard GAN architecture.
\end{itemize}

This paper is organized as follows.
In \cref{sec:related} we present two works related to our approach. In \cref{sec:sgan} we first introduce the main concepts underlying the GAN framework, then we present the main contribution of this paper: the Semantic Generative Adversarial Network architecture. The evaluation metrics used in our tests are described in \cref{sec:evaluation}. \cref{sec:experiments} contains the experimental evaluation of SemGAN. In \cref{sec:application} we describe an interesting application of this approach. Finally, \cref{sec:conclusions} contains the conclusions and future research directions.

\section{Related Work}
\label{sec:related}

\begin{figure*}[htb]
	\includegraphics[width=1\textwidth]{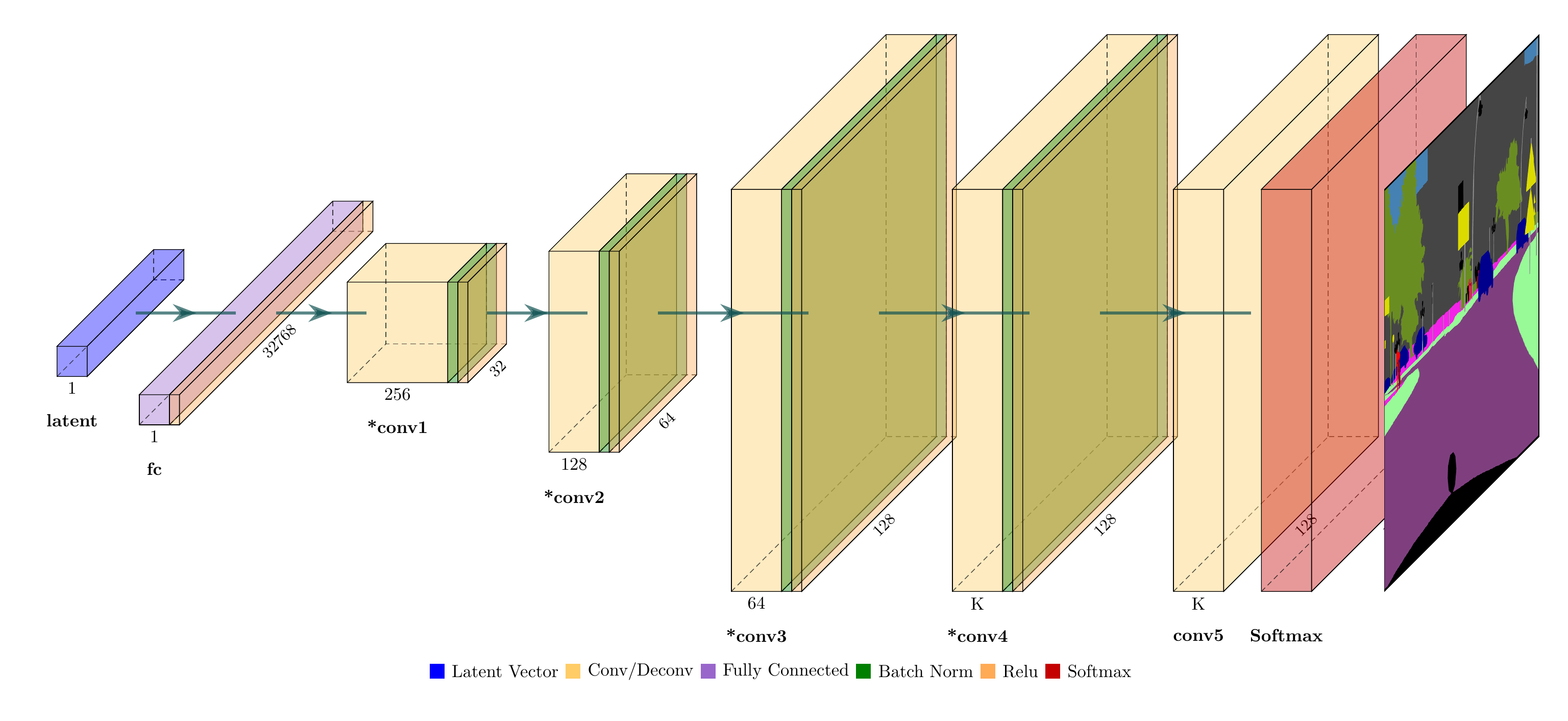}
	\caption{Illustration of semantic generator architecture. The input is a latent vector and the output is a semantic mask.}
	\label{fig:sgan-architecture}
\end{figure*}

\paragraph{GAN-based Semantic Segmentation} The generation of semantic masks can be related to \citet{luc-semantic-2016}.  Semantic-segmentation using GAN is an adversarial approach to the problem of semantic-segmentation, that is detecting for each pixel the class it belongs to. In \citet{luc-semantic-2016} the generator outputs a semantic map starting from a realistic image. The generator output map is then multiplied channel-wise by the input image before being fed into the discriminator to avoid the simple detection of distribution inconsistencies between the ground truth annotations (one-hot encoded) and the generator output. However, we highlight the fact that this approach does not allow to generate semantic images starting from a latent vector, like in unconditional generative models, therefore this model is not suitable for the generation of new artificial semantic environments. %

\paragraph{Image to Image translation}
In \citet{isolaImagetoImageTranslationConditional2016} an adversarial approach to Image to Image translation is presented (pix2pix).
The \citet{isolaImagetoImageTranslationConditional2016} approach has been applied to translate semantic images to realistic images and vice versa. When used to produce semantic labels, the output must be post-processed since the generator is not able to generate labels (colors) that perfectly match the space of semantic labels. Like in \citet{luc-semantic-2016}, the pix2pix model does not allow to generate entirely new semantic images, always requiring a source domain.

\section{Pixel-level Generative Adversarial Networks}
\label{sec:sgan}
\subsection{Generative Adversarial Networks}

Generative Adversarial Networks are based on a game-theoretic formulation in which there are two main components, namely the \textit{generator} and  the \textit{discriminator}, competing against each other in a \textit{minimax} game. Given the data distribution $p_{\text{data}}(\cdot)$ over variables $\mathbf{x}$, the goal of the generator $G$ is to learn a mapping from a prior distribution $p_g(\cdot)$ over latent variables $\mathbf{z}$ to $p_{\text{data}}(\cdot)$. On the other side, the discriminator $D$ must learn to determine whether a sample comes from the data distribution or the model distribution. We denote with $\bm{\theta}_G$ the generator parameters and with $\bm{\theta}_D$ the discriminator parameters.
Both $G$ and $D$ are trained simultaneously considering the two player minimax game:
\begin{equation}
	\begin{split}
		\min_G \max_D V(D,G) = &\mathbb{E}_{\mathbf{x} \sim p_{\text{data}}(\cdot)} \left[ \log(D(\mathbf{x}) \right] +\\
		& \mathbb{E}_{\mathbf{z} \sim p_g(\cdot)} \left[ \log  \left( 1 - D(G(\mathbf{z})) \right) \right].
	\end{split}
\end{equation}
The goal of GAN training is to find a Nash equilibrium, that is a point $(\bm{\theta}^{*}_G, \bm{\theta}^{*}_D)$ in the parameter space in which neither the generator nor the discriminator can improve their cost function unilaterally. Training GANs is well known for being unstable and prone to divergence \cite{arjovsky-towards-2017, mescheder-which-2018}. \newline
Feature matching loss \cite{salimans-improved-2016} is known to reduce the instability of GAN training. Instead of updating $G$ parameters based on the output of the discriminator, it updates $G$ based on the internal representation of $D$, e.g., the features of an intermediate layer. \newline

\subsection{Semantic Generative Adversarial Networks}

\begin{figure*}[htb]
	\includegraphics[width=1\textwidth]{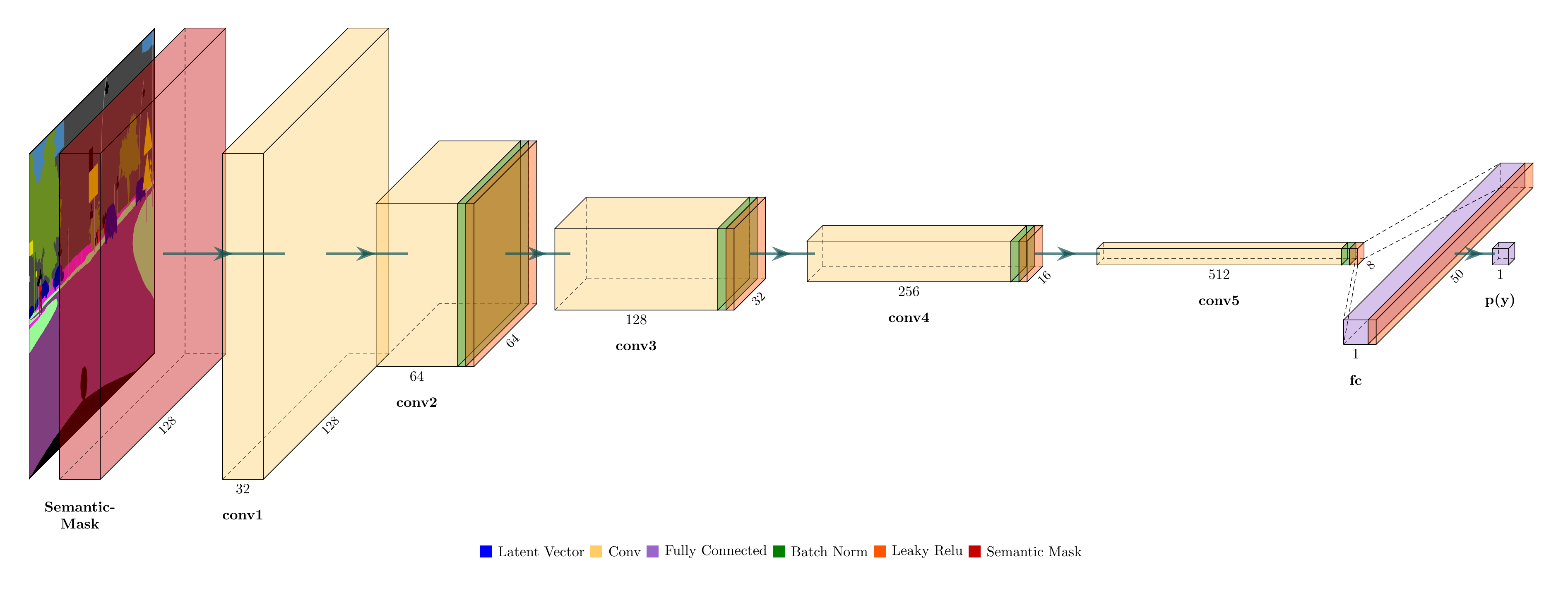}
	\caption{Illustration of semantic discriminator architecture. The input is a semantic mask and the output is the probability of being a real semantic mask.}
	\label{fig:sgan-dis-architecture}
\end{figure*}

We propose a new technique for generating semantic labels at a pixel-wise level based on Generative Adversarial Networks. The SemGAN generator outputs a tensor with shape $(W,H,K)$, where $K$ is the number of classes defined in the dataset; we denote with the name \textit{class channel} the last dimension of the generator output.
The last layer of the generator is a softmax over the class channel applied in order to interpret every pixel vector as a probability distribution over the classes.
The SemGAN discriminator input has the same shape as the output of the generator, and it corresponds to the one hot notation of the semantic map in the case of real images and to the softmax distribution over classes in the case of generated images. The SemGAN architecture is illustrated in \cref{fig:sgan-architecture} and in \cref{fig:sgan-dis-architecture}. Our architecture is a more principled way to generate semantic map rather than treating semantic maps as normal images.

The generator tends to generate distributions with very low variance since it tries to replicate the original images, that are one hot encoded. Using this approach, we are sure to generate only valid labels; moreover, we help the generator in its task because of the restricted subset to learn. Thus, the generator function $G$ can be formalized as:
\begin{equation}
	\begin{split}
		G: \mathcal{X} \rightarrow \Delta(L)^{W \times H},
	\end{split}
\end{equation}

where $\mathcal{X}$ is the latent space and with $\Delta(L)$ we denote the probability distribution induced by the softmax operation over the set of labels $L$:
\begin{equation}
	\begin{split}
		L = \{l_1, l_2, \dots, l_K\}.
	\end{split}
\end{equation}

A prediction can be collapsed into a segmentation map by taking the $\argmax$ of each depth-wise pixel vector and after mapping the labels in $L$ to their corresponding colors obtaining an RGB image naturally follows.

This architecture is very general, and there is the possibility to combine it with all defined losses as well as all training strategies available.
We highlight three important advantages of our architecture:
\begin{itemize}
    \item the generator always outputs a valid class label;
    \item the gradient signal arriving at the generator is applied to the predicted class labels, not to the generated colors;
    \item the limited generator co-domain speeds up the convergence: we empirically demonstrated to achieve better qualitative and quantitative results with fewer training steps w.r.t. the standard GAN architecture.
\end{itemize}

\section{Evaluation Metrics}
\label{sec:evaluation}
\begin{figure*}[!htb]
\centering
\bgroup 
 \setlength\tabcolsep{0.2pt}
\begin{tabular}{c c}
\hspace{-1.cm} GAN & \hspace{-2.cm} \textbf{SemGAN (ours)} \\ 
\hspace{1cm} \includegraphics[width=0.4\linewidth]{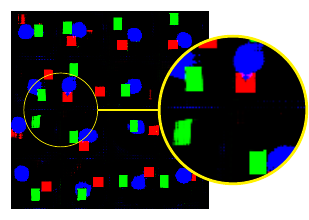}  \hspace{1cm}& \includegraphics[width=0.4\linewidth]{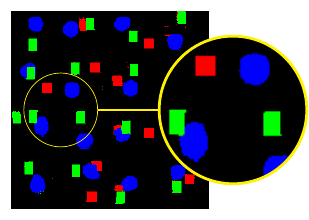}  \\
\end{tabular} \egroup 

\vspace{-0.1in}
\caption{Visual comparison for the colored shapes experiment. The images are composed by 16 figures side by side. The output of the GAN contains many artifacts while the output of the Semantic GAN contains a limited number of spourius pixels.}
\label{fig:coloroured_shapes_imgs}
\end{figure*}

Assessing image quality is always a challenging task in all GAN-based approaches. We evaluate our architecture using three metrics based on the assumption that the final goal of the generator is to generate images visually similar at every scale to the images contained in the training set. 
\paragraph{Sliced Wasserstein Distance} The first metric we use in our evaluation is the Sliced Wasserstein Similarity (SWD), used in \citet{progan}. Following this approach we build a Laplacian pyramid downsampling the considered image, we extract patches of $7 \times 7$, with three color channels. A small SWD of a certain pyramid level indicates that the generated images are structurally similar to the training set at this level.
\paragraph{Fr\'echet Inception Distance} The second metric we use is the Fr\'echet Inception Distance (FID) \cite{heuselGANsTrainedTwo2017}. This metric is based on the Fr\'echet Distance between features extracted on a particular layer of the GoogLeNet \cite{szegedyGoingDeeperConvolutions2015}. Even if this net has been trained on real images, a visual inspection shows that the particular features we are extracting might be also interesting for semantic images. As for the SWD, a small FID indicates that features of the generated images are similar to the features of the training images. 
\paragraph{MultiScale Structural Similarity} The third metric is the MultiScale Structural Similarity \citep{wangMultiscaleStructuralSimilarity2003} (MS-SIM). This metric detects the similarity among generated images, and this is useful in order to detect mode collapse, a typical GAN problem. The downside of this metric is that it does not take into account the similarity between the generator output and the training set.

\section{Experimental Evaluation}\label{sec:experiments}

\begin{figure*}[htb]
	\centering
	\includegraphics[width=0.95\textwidth]{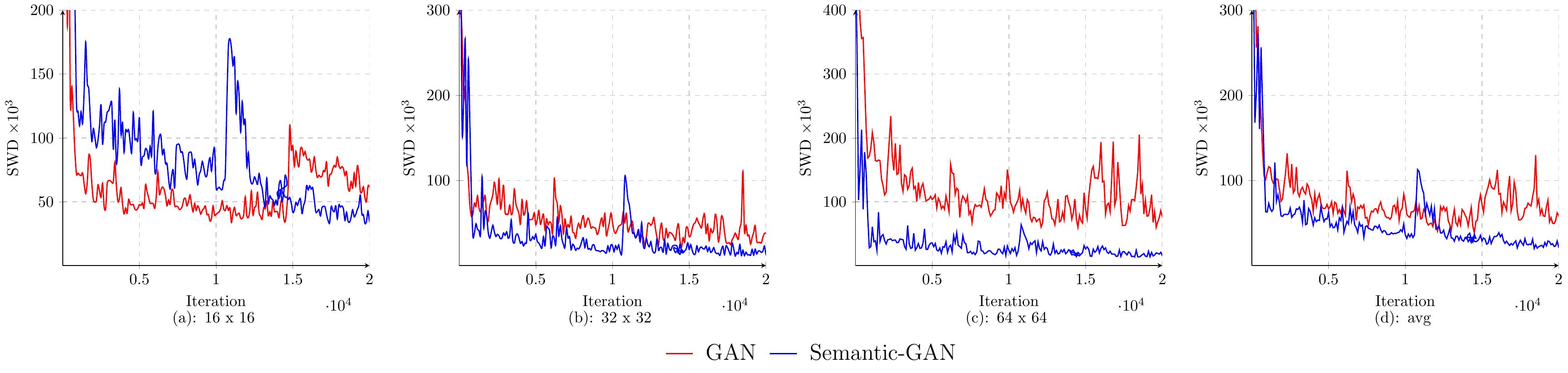}
	\caption{Sliced Wasserstein distance metric for the Coloured shapes experiment. GAN vs Semantic-GAN comparison.}
	\label{fig:coloroured_shapes_exp}
\end{figure*}

In this section, we show the experimental evaluation of our architecture on a toy dataset of colored shapes (\cref{sec:toydataset}) and two standard semantic datasets, namely Cityscapes (\cref{sec:cityscape}) and Facades (\cref{sec:facades}). We compare our results with the results obtained with a traditional architecture working on RGB images. We highlight the fact that we use the same parameters, the same loss, and the same architecture both for the semantic GAN and for the standard GAN. The unique difference is the output layer, that in the case of the Semantic GAN is a volume with dimension $W \times H \times K$ representing a probability distribution over labels for each pixel vector, while in the case of the GAN is a standard RGB image, thus with dimensions $W \times H \times 3$.

\subsubsection{Semantic Generation of Colored Shapes}
\label{sec:toydataset}
The Colored Shapes dataset takes inspiration from \citet{liuIntriguingFailingConvolutional2018}. It contains 2916 images $64\times64\times3$ each. Each image contains a $10\times10$ px red square in a different location ($54^2$ available positions), a blue circle with radius $10$ randomly placed in the image and a green rectangle in a random position. The three shapes can overlap, and the circle can completely cover the square. Thus, the classes in this dataset are 4: background, circle, rectangle, square.

We evaluate the image quality visually (\cref{fig:coloroured_shapes_imgs}) and using the SWD metric (\cref{fig:coloroured_shapes_exp}). The semantic GAN architecture outperforms the GAN architecture, especially for the SWD $64 \times 64$ metric. When observing the generated images,   artifacts can be spotted in the GAN output, while the proposed Semantic GAN output is very precise.

\subsubsection{Semantic Generation of Street Scenes}
\label{sec:cityscape}

\begin{table}[tb]
	\centering
	\setlength\tabcolsep{3pt} 
	\begin{tabular}{c|cccccccc}
	\multirow{2}{*}{} & \multirow{2}{*}{FID}           & \multirow{2}{*}{MS-SIM}      & \multicolumn{5}{c}{SWD}      \\
		& & & 16 & 32 & 64 & 128 & avg \\
		\hline
		SemGAN  & \bf{50.89} & \bf{0.1972} & \bf{20.9} & \bf{14.4} & \bf{18.4} & \bf{15.0} & \bf{20.5} \\
		GAN          & 164.3      & 0.3164     & 47.6      & 20.0      & 20.6      & 15.6      & 24.5    \\
	\end{tabular}
	\caption{Performance comparison between SemGAN (ours) and GAN for the cityscapes experiment. Best results obtained among a set of experiments and during training. For every metric lower is better.}
	\label{tab:metrics_cityscapes_main}
\end{table}

\begin{figure*}[htb]
	\centering
	\includegraphics[width=0.95\textwidth]{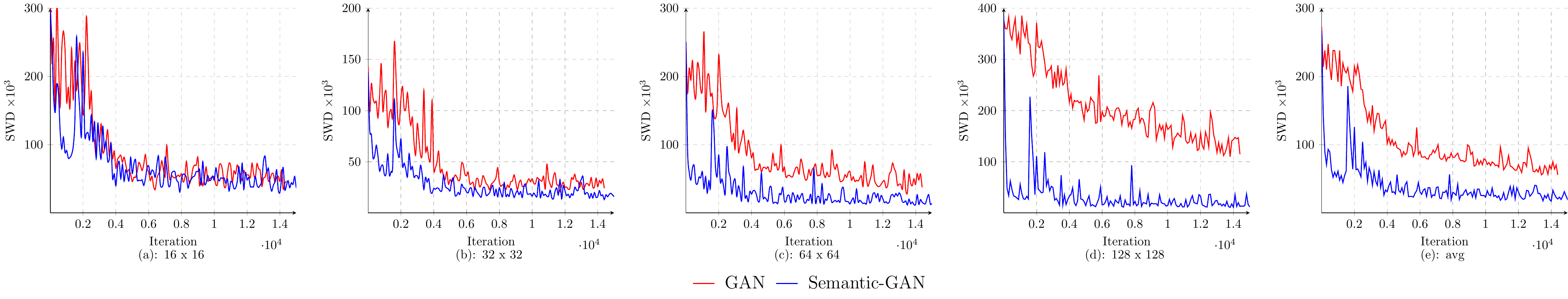}
	\caption{Sliced Wasserstein distance metric for the cityscape experiment. GAN vs Semantic-GAN comparison.}
	\label{fig:cityscape_exp}
\end{figure*}

\begin{figure*}[htb]
\centering
\bgroup 
 \def\arraystretch{1.1} 
 \setlength\tabcolsep{0.2pt}
\begin{tabular}{c c}
	 GAN & \textbf{SemGAN (ours)}  \\ \includegraphics[width=0.4\textwidth]{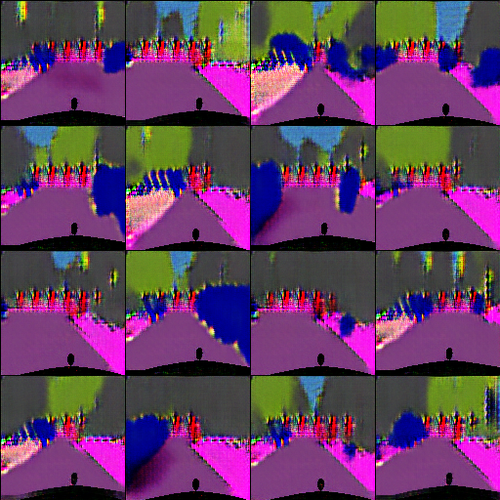}
	\hspace{0.5cm} & \includegraphics[width=0.4\textwidth]{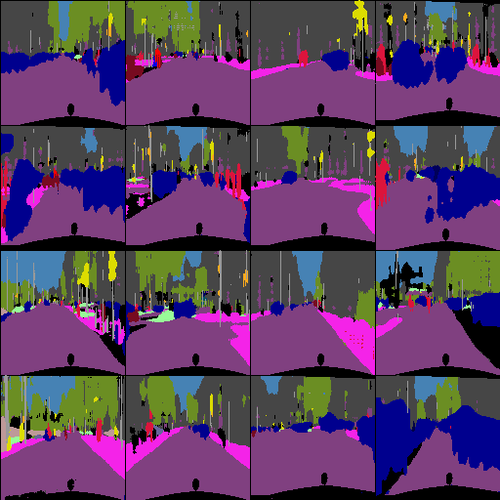}
	\\
\end{tabular} \egroup
	\caption{Visual comparison of GAN (left) and SemGAN (right) for the cityscapes experiment. Models yielding the best FID. The figures are composed by 16 images side by side.}
	\label{fig:cityscape_imgs}
\end{figure*}

The Cityscapes \citep{cordtsCityscapesDatasetSemantic2016} dataset contains a diverse set of stereo video sequences recorded in street scenes from 50 different cities. We resized the images to $128\times128$, and we used the 22 main classes. The comparison between SemGAN and GAN for the SWD metric is reported in \cref{fig:cityscape_exp}. In the lowest level of the pyramid the Sliced Wasserstein distance of the SemGAN is comparable to the one of the standard GAN, while in the highest, where the image resolution is greater, the gap increases notably. Considering the mean SWD, the SemGAN model outperforms the GAN model. \cref{fig:cityscape_imgs} depicts the generated images for each architecture. From a visual point of view, the generated images using the semantic architecture are precise, without artifacts of any sort. The generated images using the standard architecture have colors not related to any label. We also compare in \cref{tab:metrics_cityscapes_main} the FID, the MS-SIM and the SWD of the image generated by our architecture with the images generated using the standard GAN architecture. Further details on the experiments are reported in \cref{app:cityscapes}.

\subsubsection{Semantic Generation of Facades}
\label{sec:facades}
\begin{figure*}[htb]
\centering
\bgroup 
 \def\arraystretch{0.2} 
 \setlength\tabcolsep{0.2pt}
\begin{tabular}{c c}
\hspace{-1cm} GAN & \hspace{-2cm} \textbf{SemGAN (ours)}  \\ 
\hspace{1cm} \includegraphics[width=0.45\linewidth]{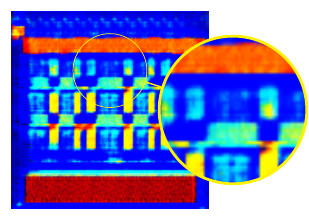}   \hspace{1cm}& \includegraphics[width=0.45\linewidth]{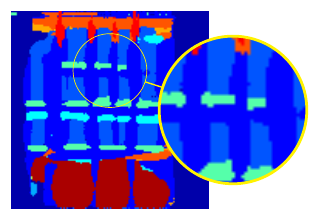} 
  \\
\end{tabular} \egroup 

\vspace{-0.1in}
\caption{Visual comparison of GAN (left) and SemGAN (right) for the facades experiment. Models yielding the best FID. The image from SemGAN is defined even if geometric shapes are not well reproduced. The image from GAN has pixels not related to any label.}
\label{fig:facades}
\end{figure*}

The Facades dataset \citep{Tylecek13} includes 606 rectified images of facades from different cities around the world and diverse architectural styles; it contains 12 classes. We resized the images to 128x128 in order to perform this experiment.
In \cref{tab:metrics_facades_main} we report the metrics used for evaluation. See \cref{fig:facades} for a visual comparison of the generated images for the best models for each architecture.
From a qualitative point of view, the generated images using SemGAN and GAN are not very impressive, this might relate to the fact that the Facades dataset has a smaller number of images with higher variance than the Cityscape dataset. We can see from the generated samples that our architecture is not able to generate precise geometric forms, unlike in the Colored Shapes experiment. Samples from the GAN are very blurry and the class distinction is not neat. Further details and experiments are reported in \cref{app:facades}.

\begin{table}[tb]
	\centering
	\setlength\tabcolsep{2.5pt}
	\begin{tabular}{c|c c c c c c c c}
	\multirow{2}{*}{} & \multirow{2}{*}{FID} & \multirow{2}{*}{MS-SIM}      & \multicolumn{5}{c}{SWD}      \\
		& & & 16 & 32 & 64 & 128 & avg \\
		\hline
		SemGAN & \bf{134.83} & \bf{0.0492} & \bf{19.8} & \bf{16.5} & \bf{21.7} & \bf{25.1} & \bf{28.4} \\
		GAN          & 198.05      & 0.0698      & 30.6      & 19.1      & 22.2      & 68.4      & 39.3      \\
	\end{tabular}
	\caption{Performance comparison between SemGAN (ours) and GAN for the facades experiment. Best results obtained among a set of experiments and during training. For every metric lower is better.}
	\label{tab:metrics_facades_main}
\end{table}

\section{Application}
\label{sec:application}
\subsection{Dreaming realistic Street Scenes}
\begin{figure*}
\centering
\bgroup 
 \def\arraystretch{0.2} 
\begin{tabular}{cccc}
SemGAN-Output & pix2pix & GAN-Output & pix2pix\\ 
\includegraphics[width=0.2\linewidth]{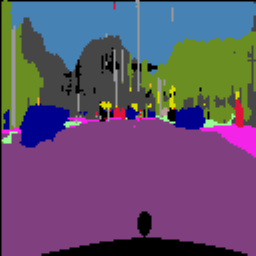} &
\includegraphics[width=0.2\linewidth]{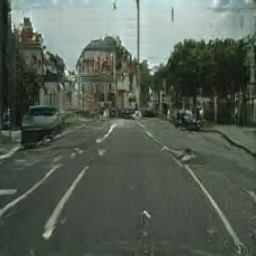} &
\includegraphics[width=0.2\linewidth]{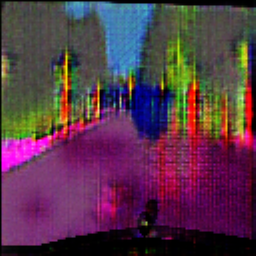} &
\includegraphics[width=0.2\linewidth]{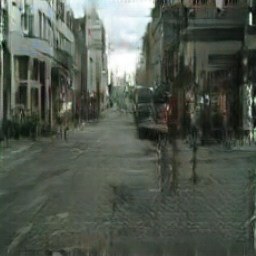} \\ 
\includegraphics[width=0.2\linewidth]{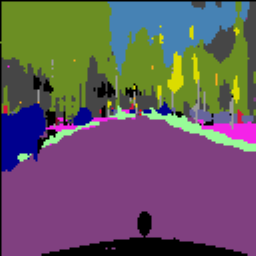} &
\includegraphics[width=0.2\linewidth]{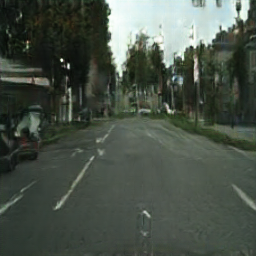} &
\includegraphics[width=0.2\linewidth]{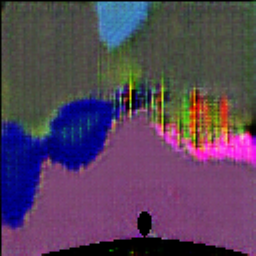} &
\includegraphics[width=0.2\linewidth]{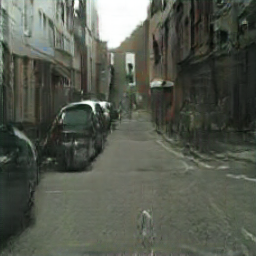} \\ 
\includegraphics[width=0.2\linewidth]{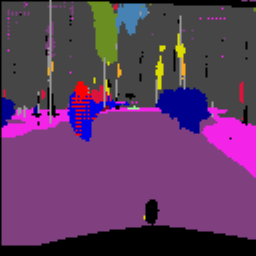} &
\includegraphics[width=0.2\linewidth]{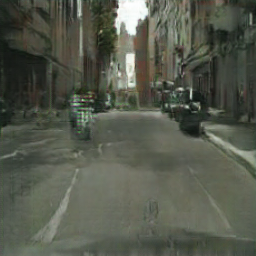} &
\includegraphics[width=0.2\linewidth]{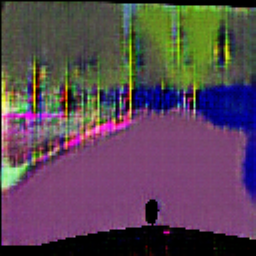} &
\includegraphics[width=0.2\linewidth]{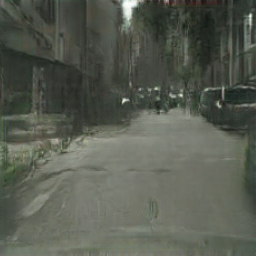}\\ 
\includegraphics[width=0.2\linewidth]{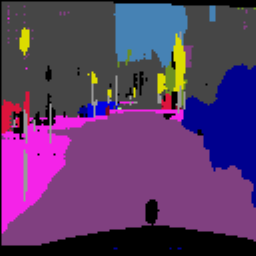} &
\includegraphics[width=0.2\linewidth]{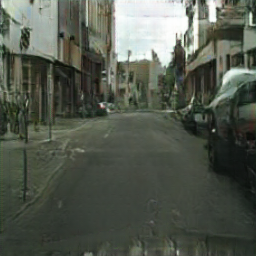} &
\includegraphics[width=0.2\linewidth]{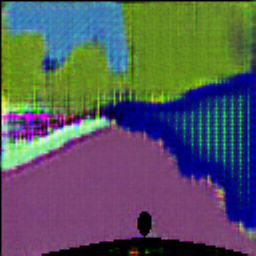} &
\includegraphics[width=0.2\linewidth]{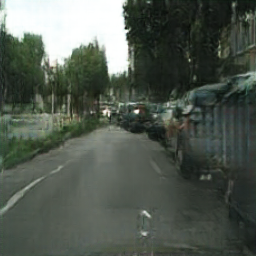}\\ 
\includegraphics[width=0.2\linewidth]{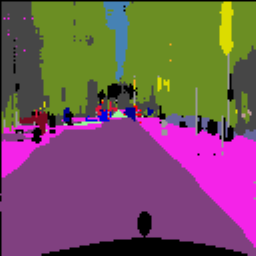} &
\includegraphics[width=0.2\linewidth]{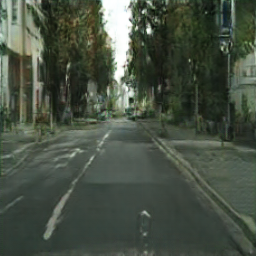} &
\includegraphics[width=0.2\linewidth]{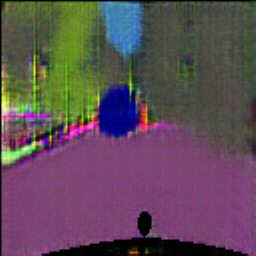} &
\includegraphics[width=0.2\linewidth]{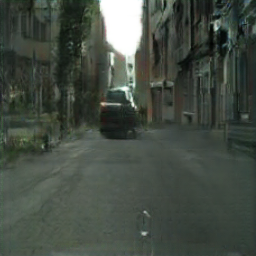}
\end{tabular} \egroup 

\vspace{-0.1in}
\caption{Comparison between the application of pix2pix to the output of SemGAN (two columns on the left) and the output of GAN (two columns on the right).}
\label{fig:sgan-p2p}
\end{figure*}

The Semantic-GAN architecture can be combined with pix2pix in order to generate new, unseen street scenes; this is obtained by feeding the output of our generator into the pix2pix generator, after having applied the RGB mapping and scaling to match the input size. Results of the pix2pix model applied to the SemGAN and GAN outputs are depicted in \cref{fig:sgan-p2p}.
The street scenes dreamed by the semantic GAN are much more realistic w.r.t. the street scenes dreamed by the standard GAN.
It is worth noting that the pix2pix model we used for generating images is trained on compressed RGB images, not on semantic maps. A pix2pix model trained on semantic maps can benefit from this representation.

\section{Conclusions}
\label{sec:conclusions}
In this paper, we presented Semantic-GAN, a new GAN architecture able to generate semantic images starting from a latent representation. The empirical evaluation suggests that using this type of architecture can be beneficial when working in semantic domains, where a pixel level accuracy is needed. Our approach outperforms the standard GAN architecture in the generation of images with pixels value in a finite domain since each pixel is modeled as a value sampled from a probability distribution induced over a finite set of discrete values.
While requiring to change only the generator output and the discriminator input, the Semantic-GAN approach can be applied to many different applications such as the learning of a semantic encoder able to map semantic maps to their latent representations. This idea paves the way for new applications like the generation of realistic worlds via latent space interpolation combining our method with a style transfer solution.

\clearpage

\bibliographystyle{icml2019}

\bibliography{bib}

\clearpage

\onecolumn
\appendix

\section{Further Results}

\subsection{Cityscapes}
\label{app:cityscapes}

\begin{figure}
\centering
 \setlength\tabcolsep{1cm}
\begin{tabular}{cc}
\textbf{SemGAN (ours)} & GAN \\ 
\includegraphics[width=0.3\linewidth]{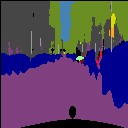} &
\includegraphics[width=0.3\linewidth]{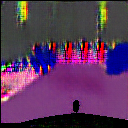} \\
\includegraphics[width=0.3\linewidth]{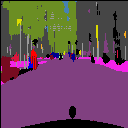} &
\includegraphics[width=0.3\linewidth]{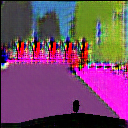}  \\
\includegraphics[width=0.3\linewidth]{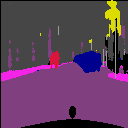} &
\includegraphics[width=0.3\linewidth]{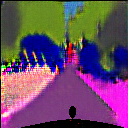}  \\
\includegraphics[width=0.3\linewidth]{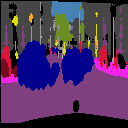} &
\includegraphics[width=0.3\linewidth]{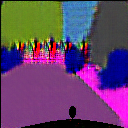}  \\
\end{tabular} 

\vspace{-0.1in}
\caption{Samples from Semantic GAN (left) and from GAN (right) trained on the cityscapes experiment. Models yielding the best FID.}
\label{fig:cityscapes-all}
\end{figure}

In this appendix we show further results for the cityscapes experiment, we provide a full comparison of SemGAN and GAN for each type of architecture we tested. For all experiments we use a batch size equal to 32.
The main parameters we focus on in our experiments are the learning rate (lr), the kernel size (ks), the latent space dimension (ld). Results are summarized in \cref{tab:metrics_cityscapes}. Sample images are reported in \cref{fig:cityscapes-all}.

\subsection{Facades}
\label{app:facades}

\begin{figure}[ht!]
\centering
 \setlength\tabcolsep{1cm}
\begin{tabular}{cc}
\textbf{SemGAN (ours)} & GAN \\ 
\includegraphics[width=0.20\linewidth]{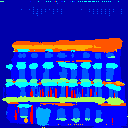} &
\includegraphics[width=0.20\linewidth]{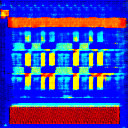} \\
\includegraphics[width=0.20\linewidth]{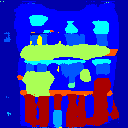} &
\includegraphics[width=0.20\linewidth]{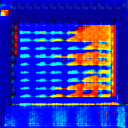} \\
\includegraphics[width=0.20\linewidth]{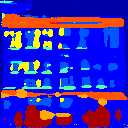} &
\includegraphics[width=0.20\linewidth]{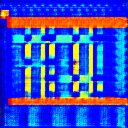} \\
\includegraphics[width=0.20\linewidth]{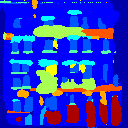} &
\includegraphics[width=0.20\linewidth]{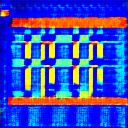} \\
\includegraphics[width=0.20\linewidth]{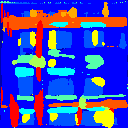} &
\includegraphics[width=0.20\linewidth]{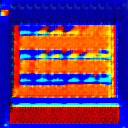}
\end{tabular} 
\caption{Samples from Semantic GAN (left) and from GAN (right) trained on the facades experiment. Models yielding the best FID.}
\label{fig:facasades-all}
\end{figure}

In this appendix we show further results for the facades experiment, we provide a full comparison of SemGAN and GAN for each type of architecture we tested. For all experiments we use a batch size equal to 32.
The main parameters we focused on in our experiments are the learning rate (lr), the kernel size (ks), the latent space dimension (ld). Results are summarized in \cref{tab:metrics_facades}. Sample images are reported in \cref{fig:facasades-all}.

\begin{table}[b!]
	\centering
	\caption{Cityscapes experiment}
	\begin{tabular}{c | c c c | *{7}{c}}
		\multirow{2}{*}{Model} & \multicolumn{3}{c|}{Architecture}  & \multicolumn{7}{c}{Metrics}                                                                                                                         \\
		\cline{2-11}
		                       & lr                                 & ld                          & ks                  & FID         & MS-SIM      & SWD 16      & SWD 32      & SWD 64      & SWD 128     & SWD avg     \\
		\hline
		SemGAN                  & \multirow{2}{*}{$2 \cdot 10^{-4}$} & \multirow{2}{*}{128}        & \multirow{2}{*}{7} & \bf{50.891}      & \bf{0.1972}      & \bf{20.9}      & \bf{14.4} & \bf{18.4}      & \bf{15.0}      & \bf{20.5}      \\
		GAN                    &                                    &                             &                     & 164.33   & 0.3164     & 47.6     & 21.7     & 61.0    & 245.1     & 96.1      \\
		\hline
		SemGAN                  & \multirow{2}{*}{$2 \cdot 10^{-4}$} & \multirow{2}{*}{300}        & \multirow{2}{*}{11}  & 55.144      & 0.211 & 26.9      & 16.4     & 20.6     & 15.6      & 24.5      \\
		GAN                    &                                    &                             &                     & 258.56    & 0.5768     & 51.3      & 20.6      & 23.8     & 73.3      & 51.6      \\
		\hline
		SemGAN                  & \multirow{2}{*}{$2 \cdot 10^{-4}$} & \multirow{2}{*}{300}        & \multirow{2}{*}{7}  & 61.422 & 0.210      & 35.3 & 20.0      & 20.7 & 17.6      & 24.6 \\
		GAN                    &                                    &                             &                     & 208.09    & 104.4      & 80.4      & 33.5      & 29.8    & 68.4      & 41.4      \\
		
	\end{tabular}
	\caption*{Performance comparison between SemGAN and GAN. Best results obtained during training ($5\cdot 10^4$ steps). For every metric lower is better. For the architecture we denote with lr the learning rate, with ld the latent space dimension, with ks the kernel size used both in the generator and in the discriminator. For the metrics we denote with FID the Frechet Inception Distance, with MS-SIM the Multi-Scale Structural Similarity, with SWD $N$ the Sliced Wasserstein Distance (multiplied by $10^3$) obtained on subsampled image with size $N\times N$, with SWD avg the average SWD among different image sizes.}
	\label{tab:metrics_cityscapes}
\end{table}

\begin{table}[b!]
	\centering
	\caption{Facades experiment}
	\begin{tabular}{c | c c c | *{7}{c}}
		\multirow{2}{*}{Model} & \multicolumn{3}{c|}{Architecture}  & \multicolumn{7}{c}{Metrics}                                                                                                                         \\
		\cline{2-11}
		                       & lr                                 & ld                          & ks                  & FID         & MS-SIM      & SWD 16      & SWD 32      & SWD 64      & SWD 128     & SWD avg     \\
		\hline
		SemGAN                  & \multirow{2}{*}{$10^{-4}$}         & \multirow{2}{*}{300}        & \multirow{2}{*}{7}  & 145.80      & 0.0765      & 52.8      & 28.6      & 26.0      & \bf{25.1} & 41.4      \\
		GAN                    &                                    &                             &                     & 249.14      & 0.1282      & 72.5      & 47.5      & 40.0      & 80.1      & 63.2      \\
		\hline
		SemGAN                  & \multirow{2}{*}{$2 \cdot 10^{-4}$} & \multirow{2}{*}{128}        & \multirow{2}{*}{7} & 145.55      & 0.0578      & 29.5      & \bf{16.5} & 22.4      & 36.2      & 33.7      \\
		GAN                    &                                    &                             &                     & 211.79   & 0.0814     & 30.6      & 19.1      & 22.2     & 73.8      & 39.3      \\
		\hline
		SemGAN                  & \multirow{2}{*}{$2 \cdot 10^{-3}$} & \multirow{2}{*}{300}        & \multirow{2}{*}{7}  & 165.05      & \bf{0.0492} & 30.2      & 20.2      & 27.5      & 45.0      & 37.5      \\
		GAN                    &                                    &                             &                     & 229.85    & 0.0698     & 51.3      & 20.6      & 23.8      & 73.3      & 51.6      \\
		\hline
		SemGAN                  & \multirow{2}{*}{$2 \cdot 10^{-4}$} & \multirow{2}{*}{300}        & \multirow{2}{*}{7}  & \bf{134.83} & 0.0557      & \bf{19.8} & 17.1      & \bf{21.7} & 35.6      & \bf{28.4} \\
		GAN                    &                                    &                             &                     & 208.09    & 0.1044      & 80.4      & 33.5      & 29.8    & 68.4      & 41.4      \\
	\end{tabular}
	\caption*{Performance comparison between SemGAN and GAN. Best results obtained during training ($5\cdot 10^4$ steps). For every metric lower is better. For the architecture we denote with lr the learning rate, with ld the latent space dimension, with ks the kernel size used both in the generator and in the discriminator. For the metrics we denote with FID the Frechet Inception Distance, with MS-SIM the Multi-Scale Structural Similarity, with SWD $N$ the Sliced Wasserstein Distance (multiplied by $10^3$) obtained on subsampled image with size $N\times N$, with SWD avg the average SWD among different image sizes.}
	\label{tab:metrics_facades}
\end{table}

\section{Training details}
For each experiment we train the generator with feature matching loss and the discriminator with the binary cross entropy loss. The SWD and the FID are measured over the whole set of images in the training set. We run each training for $5 \times 10^4$ steps.
Each component of the latent vector comes from the distribution $\mathcal{N}(0,1)$.
We use the same learning rate both for the generator and the discriminator using the Adam optimizer, with $\beta_1 = 0.5$, $\beta_2=0.999$ and $\epsilon=10^{-8}$.
\paragraph{Colored shapes experiment}
For the colored shapes experiment we use a batch size equal to 32, $lr=10^{-4}$, $ks=11$, $ld=100$.


\end{document}